%% file: main.tex
\documentclass[runningheads]{llncs}
\usepackage{graphicx}
\usepackage[T1]{fontenc}
\usepackage{graphicx}
\usepackage{graphicx}
\usepackage{algorithmic}
\usepackage{amssymb}
\usepackage{algorithm}
\usepackage{color}
\usepackage{colortbl}
\usepackage{siunitx}
\usepackage{booktabs}
\usepackage{threeparttable}
\usepackage{multirow}
\usepackage{marvosym}
\usepackage{amsmath}
\usepackage{bm}
\usepackage[colorlinks,hyperindex,breaklinks]{hyperref}
\usepackage{amssymb} 
\usepackage{pifont} 
\usepackage{arydshln}
\definecolor{anti-flashwhite}{rgb}{0.95, 0.95, 0.96}
\definecolor{whitesmoke}{rgb}{0.94, 0.94, 0.94}
\definecolor{teagreen}{rgb}{0.82, 0.94, 0.75}
\definecolor{powderblue}{rgb}{0.69, 0.88, 0.9}
\definecolor{pastelblue}{rgb}{0.68, 0.78, 0.81}
\definecolor{lightskyblue}{rgb}{0.53, 0.81, 0.98}
\definecolor{turquoise}{cmyk}{0.65,0,0.1,0.3}
\definecolor{purple}{rgb}{0.65,0,0.65}
\definecolor{dark_green}{rgb}{0, 0.5, 0}
\definecolor{orange}{rgb}{0.8, 0.6, 0.2}
\definecolor{red}{rgb}{0.8, 0.2, 0.2}
\definecolor{darkred}{rgb}{0.6, 0.1, 0.05}
\definecolor{blueish}{rgb}{0.0, 0.3, .6}
\definecolor{light_gray}{rgb}{0.7, 0.7, .7}
\definecolor{pink}{rgb}{1, 0, 1}
\definecolor{greyblue}{rgb}{0.25, 0.25, 1}
\definecolor{light_cyan}{rgb}{0.88,1,1}
\definecolor{freeze}{RGB}{102,162,216}
\definecolor{train}{RGB}{211,97,3}

\newcommand{\ie}{\textit{i}.\textit{e}.}
\newcommand{\eg}{\textit{e}.\textit{g}.}
\newcommand{\etal}{\textit{et al}.}

\makeatletter

\newcommand{\Rmnum}[1]{\expandafter\@slowromancap\romannumeral #1@}
\makeatother

\begin{document}
\title{Exploring Visual Prompts for Whole Slide Image Classification with Multiple Instance Learning}
\author{Yi Lin$^{\dag}$\inst{1} \and Zhongchen Zhao$^{\dag}$\inst{2} \and Zhengjie ZHU\inst{1} \and Lisheng Wang\textsuperscript{\Letter}\inst{2} \and Kwang-Ting Cheng\inst{1} \and Hao Chen\textsuperscript{\Letter}\inst{1}}
\institute{
    The Hong Kong University of Science and Technology, Hong Kong, China.
    \and Shanghai Jiao Tong University, China. \\
    \email{\email{lswang@sjtu.edu.cn}};
    \email{jhc@cse.ust.hk} 
}
\titlerunning{Exploring Visual Prompts for MIL-based WSI Classification.}
\authorrunning{Lin~\etal}
\maketitle 
\def\thefootnote{$\dag$}\footnotetext{Equal contribution; \Letter~corresponding author.}
\input{Section/abstract}
\input{Section/introduction}

\input{Section/related_work}
\input{Section/method}

\input{Section/experiment}
\input{Section/conclusion}
\bibliographystyle{splncs04}
\bibliography{ref.bib}
\end{document}

%% file: Section/abstract.tex
\begin{abstract}
Multiple instance learning (MIL) has emerged as a popular method for classifying histopathology whole slide images (WSIs). 
However, existing approaches typically rely on pre-trained models from large natural image datasets, such as ImageNet, to generate instance features, which can be sub-optimal due to the significant differences between natural images and histopathology images that lead to a domain shift.
In this paper, we present a novel, simple yet effective method for learning domain-specific knowledge transformation from pre-trained models to histopathology images. 
Our approach entails using a prompt component to assist the pre-trained model in discerning differences between the pre-trained dataset and the target histopathology dataset, resulting in improved performance of MIL models.
We validate our method on two publicly available datasets, Camelyon16 and TCGA-NSCLC. 
Extensive experimental results demonstrate the significant performance improvement of our method for different MIL models and backbones.
Upon publication of this paper, we will release the source code for our method.
\keywords{Visual prompt \and Multiple instance learning \and Whole slide image \and Deep learning.}
\end{abstract}

%% file: Section/introduction.tex
\section{Introduction}
Whole slide images (WSI) play a vital role in histopathology image analysis and clinical disease diagnosis~\cite{1,2,3}. 
With the advent of deep-learning-based techniques, histopathology image analysis has undergone a significant transformation~\cite{4,6}. However, there are still challenges when it comes to classifying WSI. Due to their massive size, WSIs cannot be directly fed into typical deep-learning models. Therefore, WSIs are often divided into patches for processing~\cite{hou2016patch}. 
Unfortunately, annotating patch-level labels is labor-intensive and time-consuming, which limits the applicability of conventional supervised learning methods~\cite{7,8}. To address this issue, multiple instance learning (MIL) has emerged as the dominant technique for WSI analysis. In this approach, each WSI is considered as a bag containing multiple patches (instances), and a WSI bag is labeled negative only if all patches (instances) of this bag are negative. Conversely, the bag's label is positive if at least one of its instances is positive. 

Downsampling and feature extraction are necessary due to the large number of patches in a WSI. 
The quality of the extracted patch features greatly influences the performance of the subsequent MIL classification. 
Most existing methods~\cite{ABMIL,DSMIL,DTFD,CLAM} extract patch features by a frozen feature extractor pre-trained on the large natural image datasets, such as ImageNet~\cite{deng2009imagenet}, and then train the MIL classifier for the WSI prediction, as shown in Fig.~\ref{fig_innovation} $(a)$.  
However, such a MIL training scheme overlooks the domain shift issue between natural and pathological images.
To narrow the domain shift, some researchers~\cite{ssl} propose to use self-supervised pre-training methods such as SimCLR~\cite{chen2020simple} to train the feature extractor. 
However, these self-supervised learning methods do not take full advantage of the bag labels, resulting in limited performance. 
Another naive solution is to use partial patches instead of all patches to fine-tune the pre-trained feature extractor, as illustrated in Fig.~\ref{fig_innovation} $(b)$. 
However, fine-tuning all the parameters of the feature extractor using limited patches without patch-level labels may impair the benefits from pre-training on large-scale datasets like ImageNet, which increases the risk of overfitting in downstream tasks. 

\input{Figure/innovation}

Inspired by the breakthrough of prompt learning in natural language processing (NLP), we introduce visual prompts to adapt the pre-trained feature extractor to pathological images, addressing the aforementioned issue.
Our prompt learning framework for MIL-based WSI classification is shown in Fig.~\ref{fig_innovation} $(c)$.
Limited by memory capacity, we propose a feasible solution that selects representative pathological images, instead of all patches, to fine-tune the pre-trained feature extractor.
Based on the selected images, we design a prompt component added to the feature extractor to learn visual prompts, and freeze the backbone while only training the prompt component with the lightweight MIL classifier.
In this way, our method can improve the performance of the ImageNet pre-trained feature extractor and achieve domain transformation to pathological image data.
Our method also makes the entire training process highly efficient and lightweight. 
To the best of our knowledge, this is the first work to explore prompt learning for WSI classification. 
In summary, our contributions are three-fold:
\begin{itemize}
    \item[$\bullet$] We, for the first time, introduce visual prompts into WSI classification, which enables data domain transformation by learning prompt components.
    \item[$\bullet$] We propose an intuitive but effective method for end-to-end prompt training, which involves representative patch selection to reduce the number of instances in a WSI bag. 
    \item[$\bullet$] We conduct extensive experiments to validate the effectiveness of the proposed method on two public datasets, \ie, Camelyon16 and TCGA-NSCLC. Experimental results demonstrate consistent improvements across different MIL methods and backbones.
\end{itemize}

%% file: Figure/innovation.tex
\begin{figure*}[t!]
	\centering
	\includegraphics[width=0.75\textwidth]{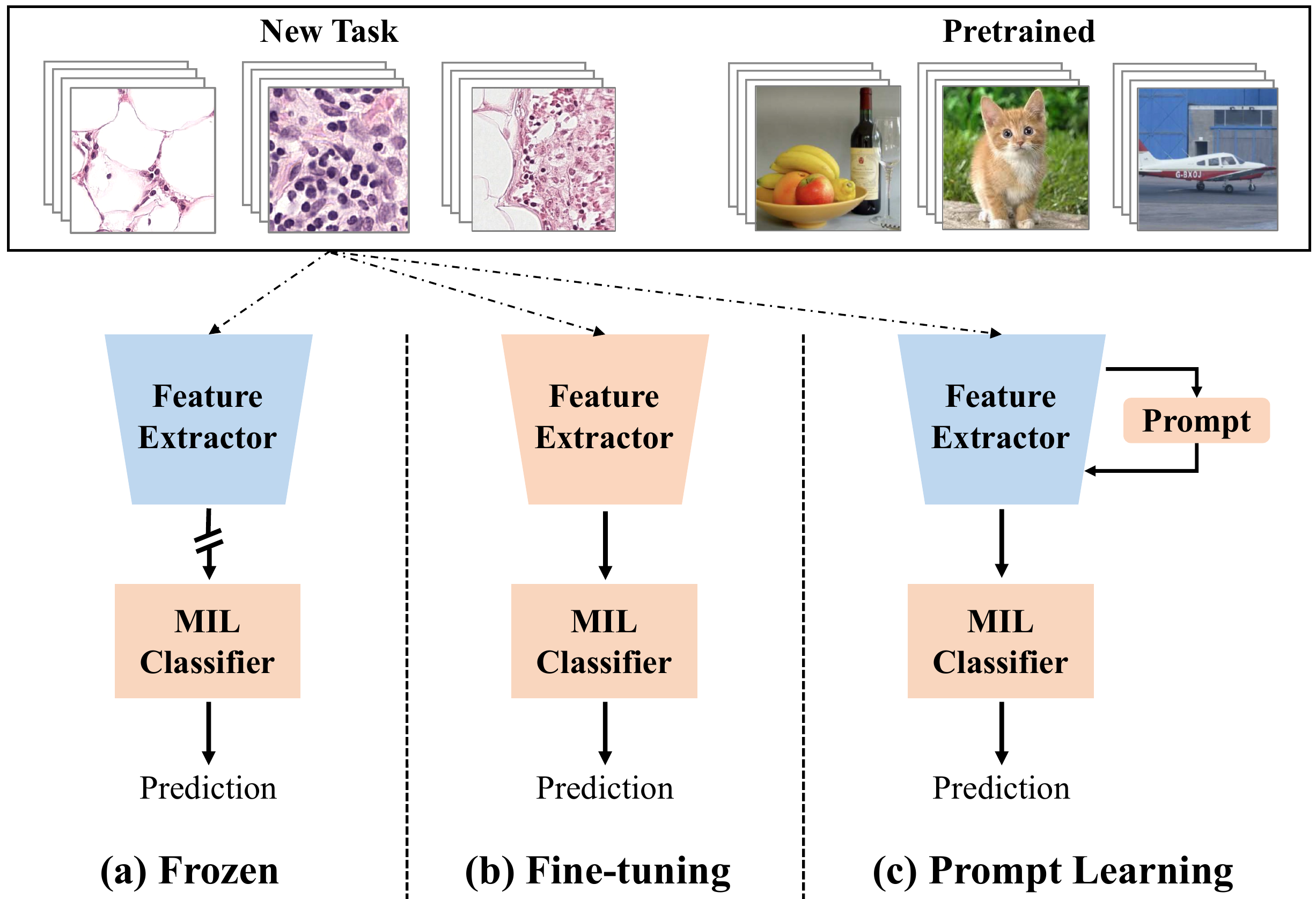}
	\caption{An illustration of three adapting schemes for WSI classification with MIL: (a) \textcolor{freeze}{\textbf{freezing}} the pretrained feature extractor while only \textcolor{train}{\textbf{training}} the MIL classifier; (b) fine-tuning the feature extractor and training the MIL classifier; (c) only training the prompt blocks and the MIL classifier.}
	\label{fig_innovation}
\end{figure*}

%% file: Section/related_work.tex
\section{Related Work}
\textbf{MIL for WSI classification.}
MIL-based methods~\cite{ABMIL,DSMIL,DTFD,CLAM} have gained popularity in WSI classification due to their high effectiveness. These methods typically involve using a feature extractor to extract features from image patches of WSI, followed by an aggregation step to obtain a feature representation at the WSI level. A lightweight classifier is then employed to predict the WSI category~\cite{ABMIL}.
In WSI classification, an effective feature extractor that generates representative feature is crucial for accurate classification results. 
However, existing MIL-based methods for WSI classification mainly adopt a pre-trained feature extractor without fine-tuning, which results in sub-optimal performance~\cite{hu2022predicting}. 
This is due to the domain shift and task discrepancies between the pre-training task (\eg, ImageNet) and the downstream task (\eg, histopathology)~\cite{stacke2020measuring}. 
For this issue, we introduce visual prompts  for WSI classification, enabling smooth feature modulation from the upstream dataset to the downstream WSI classification.

\noindent\textbf{Prompt Learning.}
Prompt learning has recently emerged as a lightweight and efficient transfer learning paradigm in NLP and has achieved remarkable success~\cite{jia2022visual}. 
The fundamental idea behind prompt learning is to freeze large-scale NLP models, such as BERT~\cite{devlin2018bert} and GPT-3~\cite{brown2020language}, that have been pre-trained on vast datasets and use task-specific prompts to adapt them to diverse downstream tasks without updating any parameters~\cite{bahng2022exploring}.
Building on the NLP prompt learning paradigm, several studies~\cite{luddecke2022image,nie2022pro,chen2022conv} have proposed to extend prompt learning to natural images in computer vision. 
For example, L{\"u}ddecke et al.~\cite{luddecke2022image} used text and image prompts to adapt the frozen pre-trained CLIP model \cite{radford2021learning} to new image segmentation tasks. 
However, the effectiveness of prompt learning in the field of histopathology analysis is under-investigated.

%% file: Section/method.tex
\section{Method}
Fig.~\ref{fig_overview} illustrates the proposed prompt learning framework, which consists of three primary steps: (\Rmnum{1}) MIL classifier training, (\Rmnum{2}) representative patch selection, and (\Rmnum{3}) prompt fine-tuning.
First, an ImageNet pre-trained ResNet~\cite{he2016deep} is used to extract patch features, which are then used to train the MIL classifier for the attention score of each patch. 
Second, representative patches in each WSI bag are chosen based on their attention scores to form a new bag. 
Third, the representative patches are used to fine-tune the prompt blocks plugged into the feature extractor and MIL classifier in an end-to-end manner.
we will elaborate on each step in the following sections.

\input{Figure/Overview} 

\subsection{Attention-based MIL Classifier with Frozen Feature Extractor}
In attention-based MIL for WSI classification, the standard training process first uses a frozen feature extractor $f(\cdot)$ pre-trained on ImageNet to extract all patch features.
Then all patch features in a WSI bag are aggregated to form the WSI feature using the attention mechanism, which learns an attention score $\alpha_{k}$ for each patch $k$ through the MIL classifier. 
The WSI feature $\bm{F}$ is obtained by computing the attention-weighted average of all patch features in a WSI as~\cite{ABMIL}: 
\begin{equation}
{\bm{F}} = \sum\limits_{k = 1}^{{K}} {{\alpha _{k}}{f(\bm{x}_{k})}},
\end{equation}  
where
\begin{equation}
{\alpha_k} = \dfrac{{\exp \left\{ {{{\bm{w}}^{\rm{T}}}\left( {\tanh \left( {{{\bm{V}}_1}{f(\bm{x}_{k})}} \right) \odot {\rm{sigmoid}}\left( {{{\bm{V}}_2}{f(\bm{x}_{k})}} \right)} \right)} \right\}}}{{\sum\limits_{j = 1}^K {\exp } \left\{ {{{\bm{w}}^{\rm{T}}}\left( {\tanh \left( {{{\bm{V}}_1}{f(\bm{x}_{k})}} \right) \odot {\rm{sigmoid}}\left( {{{\bm{V}}_2}{f(\bm{x}_{k})}} \right)} \right)} \right\}}},
\end{equation}  
where $\bm{w}$, $\bm{V}_1$ and $\bm{V}_2$ are learnable parameters in the MIL classifier, $\odot$ is the element-wise multiplication, and $\tanh(\cdot)$ and $\rm{sigmoid(\cdot)}$ denote the tanh and sigmoid activation function, respectively. \\
Finally, the MIL classifier head $h(\cdot)$ predicts the label of WSI from the WSI feature $\bm{F}$, represented as:
\begin{equation}
\tilde{\bm y} = h(\bm{F}),
\end{equation}  
where $\tilde{\bm y}$ denotes the prediction of the WSI label. 
During training, we minimize the prediction error using the cross-entropy (CE) loss.

\subsection{Representative Patch Selection.}
In WSI classification, it's common for only a small number of patches within a WSI to be associated with the disease of interest. 
For example, in the positive slides of Camelyon16~\cite{bejnordi2017diagnostic}, on average, less than 10\% of the patches in a WSI are tumor patches. 
Thus, only a few patches are sufficient to represent the entire WSI bag. 
Based on this observation, we propose a feasible solution that selects representative pathological images to fine-tune the pre-trained feature extractor for WSI classification.
In attention-based MIL, the patch with a higher attention score $\alpha_{k}$ in a WSI bag is more likely to have the same category semantics as the WSI.
Thus, based on the patches' attention score $\alpha_{k}$ calculated by the MIL classifier, we select the top-$K$ patches with the highest attention scores in each WSI as a new bag. 
Here, $K$ is set to 200, which will be discussed in Section~\ref{Implementation}. 
The new bag label is assigned as the original WSI bag label for prompt fine-tuning.
In this way, we reduce vast quantities of patches in each WSI to a small subset, enabling an end-to-end training for both feature extractor and MIL classifier.

\subsection{Prompt Fine-tuning.}
With the selected representative patches, we design a prompt learning framework to adapt the feature extractor from the natural image domain to the pathological image domain, while retaining the advantage of pre-training on the large and diverse ImageNet dataset.
Specifically, we design a prompt block that sequentially consists of a global average pooling (GAP), a multi-layer perceptron (MLP) with two layers, and a sigmoid activation, as shown in Fig.~\ref{fig_overview}. 
Given the intermediate feature map $\bm{f}_i$ from the $i$-th block of the feature extractor, the prompt block is added in parallel to the basic ResNet block $g_i(\cdot)$ to generate the visual prompt $\bm{p}_i \in {\mathbb{R}^{D}}$, where $D$ denotes the dimension of the prompt vector.
Subsequently, the generated prompts $\bm{p}_i$ are channel-wise multiplied with feature maps $\bm{f}_{i+1}$ in the next block, represented as:
\begin{equation}
\bm{p}_i = {\rm{sigmoid}}( \bm{W}_{i2} {\rm{ReLU}}(\bm{W}_{i1} {\rm{GAP}}(\bm{f}_i))),
\end{equation}
\begin{equation}
\bm{f}_{i+1} = g_i(\bm{f}_i) \odot \bm{p}_i,
\end{equation}
where $\text{ReLU}(\cdot)$ denotes rectified linear unit, $\bm{W}_{i1}$ and $\bm{W}_{i2}$ are the learnable parameters to be fine-tuned, and the parameters of $g_i(\cdot)$ remain frozen during training.
During the training process, only the parameters of prompt blocks and the lightweight MIL classifier are updated in an end-to-end manner, while the original pre-trained feature extractor is frozen.

%% file: Figure/Overview.tex
\begin{figure}[!t]
	\centering
	\includegraphics[width=1\textwidth]{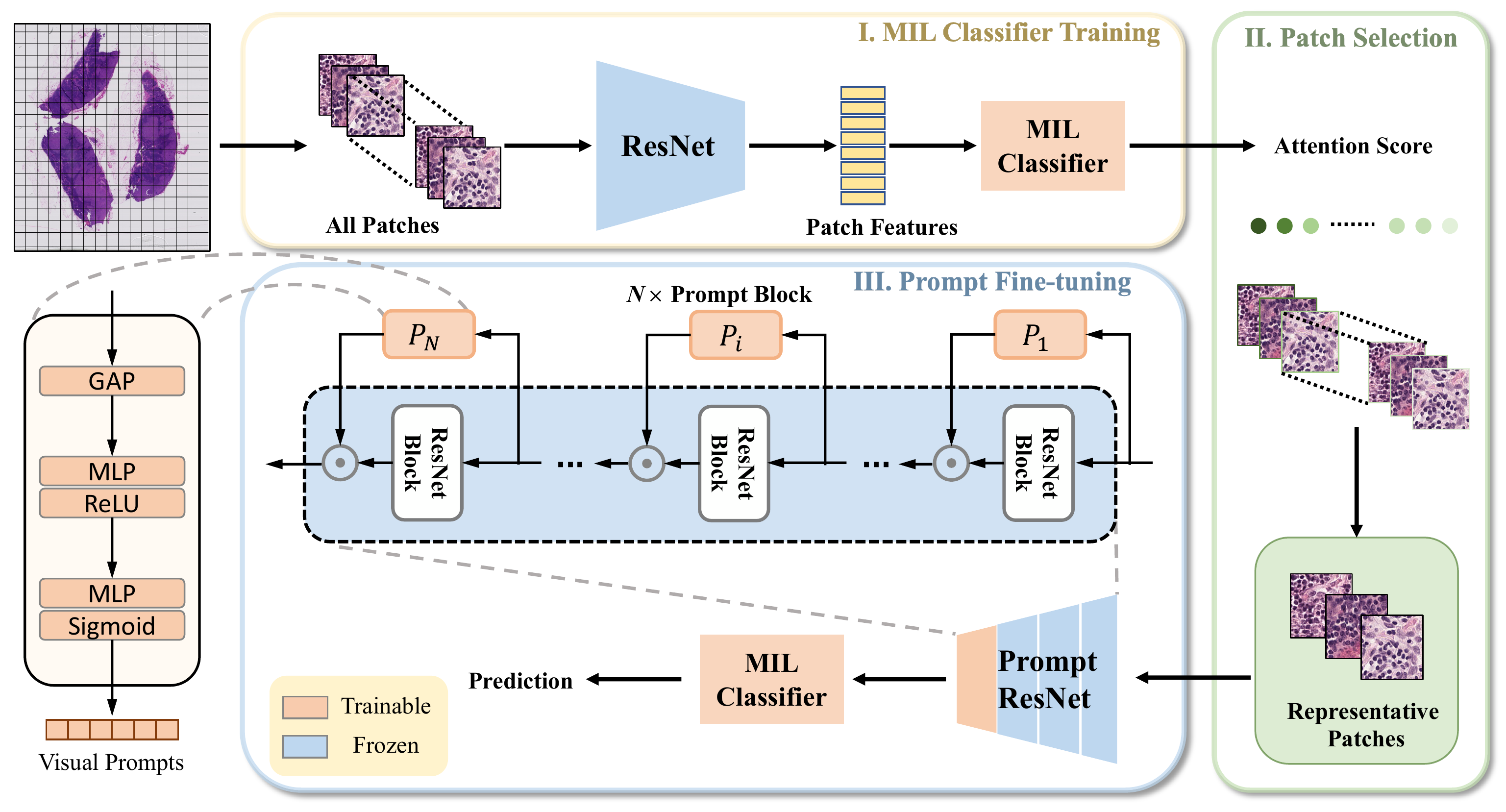}
	\caption{Overview of our method. (\Rmnum{1}) MIL classifier training with all patch features. (\Rmnum{2}) Representative patch selection. (\Rmnum{3}) Prompt learning with representative patches.}
	\label{fig_overview}
\end{figure}

%% file: Section/experiment.tex
\section{Experiments}
\subsection{Datasets}
\textbf{CAMELYON16.} 
The Camelyon16~\cite{bejnordi2017diagnostic} dataset consists of 399 H\&E stained slides from breast cancer screening, with two classes: normal and tumor. 
We employ the official 129 testing set, and the official 270 training set is further divided randomly into training and validation sets, with a ratio of $9\colon1$. After preprocessing, a total of 4,610,687 patches with the size of $256\times256$ are obtained at $20\times$ magnification, with an average of 11,556 patches per slide. \\
\textbf{TCGA-NSCLC.} 
The Cancer Genome Atlas\footnote{\url{https://www.cancer.gov/tcga}} (TCGA) non-small cell lung cancer (NSCLC)~\cite{newman2015robust} dataset is comprised of two subtypes of lung cancer: Lung Adenocarcinoma (TCGA-LUAD) and Lung Squamous Cell Carcinoma (TGCA-LUSC). It contains a total of 1,053 WSIs, with 541 LUAD slides and 512 LUSC slides. We split the dataset into training, validation, and testing sets at the slide level, with a distribution ratio of 65:10:25. In our study, we extract $256\times256$ patches at $20\times$ magnification from each WSI. After preprocessing, a total of 3,252,431 patches are obtained, with an average of 3,089 patches per WSI.
\subsection{Implementation Details}
\label{Implementation}
In our approach, we adopt the ResNet model as the feature extractor, which is pre-trained on the ImageNet dataset.
We remove the last layer of the ResNet following~\cite{CLAM} and add prompt blocks to the third layer.
For ResNet-18 and ResNet-50, we set the number of prompt blocks as 2 and 6, respectively.
For representative patch selection, we select the top 200 patches for each WSI bag by default.
For the prompt fine-tuning, we freeze the backbone of ResNet and only train the prompt block and MIL classifier using the Adam optimizer~\cite{kingma2014adam} with a learning rate of 1e-4 and weight decay of 1e-4 for 100 epochs. 
All the experiments are conducted with an NVIDIA GeForce RTX 3090 GPU. 

\subsection{Comparison Results}
\input{Table/table1}
In this study, we comprehensively evaluate the effectiveness of our proposed framework with various experiment settings, including two datasets: Cameylon16~\cite{bejnordi2017diagnostic} and TCGA-NSCLC~\cite{newman2015robust}; two backbone networks: ResNet-18 and ResNet-50; and two MIL classifiers: a state-of-the-art model DTFD~\cite{DTFD} and a common model ABMIL~\cite{ABMIL}. We report the area under curve (AUC), accuracy (Acc), and F1-score as evaluation metrics for WSI classification task.
Table~\ref{table1} shows the eight different settings of experiments. The first row in each setting represents the baseline approach with the frozen feature extractor. The second row (RPS-FT) represents using the representative patches to fine-tune both the feature extractor and MIL classifier. The third row (RPS-PT) represents our prompt fine-tuning method.
 
In Table~\ref{table1}, we can observe that our method achieves a consistent improvement compared to both the baseline approach and the fine-tuning approach.
Notably, when using ResNet-18 as the feature extractor, our method achieves a higher AUC than the DTFD baseline, with an improvement of 3.83\% on the Camelyon16 dataset and 1.28\% on the TCGA-NSCLC dataset. 
Notably, our method with prompt blocks significantly outperforms the RPS-FT scheme, demonstrating the advantage of our visual prompts.
Besides, our method shows particularly superior results on the Camelyon16 dataset compared to the results on the TCGA-NSCLC dataset. 
This can be attributed to the fact that the Camelyon16 dataset has fewer WSIs and a smaller proportion of tumor regions. 
This further confirms the advantage of our approach in handling challenging datasets. Overall, the experiment results indicate that our method effectively improves the performance of attention-based MIL methods on the WSI classification tasks. 

\subsection{Ablation Study}
\input{Figure/Ablation}
\noindent\textbf{Effectiveness of the prompt block number.}
To study the impact of the prompt component number, we conduct experiments on the Camelyon16 dataset using DTFD as the MIL classifier with different numbers of prompt blocks added to ResNet-50.
The results presented in Fig.~\ref{fig_ablation} $(a)$ show that the AUC values with different prompt quantities are consistently improved by approximately 1\% compared to the baseline approach when using more than one prompt block. 
These results suggest that our prompt block component is robust and effectively improves the performance of the feature extractor.

\noindent\textbf{Influence of the number of representative patches.}
We investigate the effect of different top-$K$ values in the RPS procedure on the performance of our method using ResNet-50 and DTFD on the Camelyon16 dataset. 
Fig.~\ref{fig_ablation} $(b)$ shows the AUC values under two different settings: fine-tuning scheme (RPS-FT) and our prompt fine-tuning scheme (RPS-PT). 
Our method consistently outperforms the fine-tuning scheme across all $K$ values using less than 50\% GPU resources, demonstrating the efficiency and effectiveness of our method.

\subsection{Analysis on the Effectiveness of Visual Prompts}
During the experiment, we found that fine-tuning produced unsatisfactory performance, which was even inferior to the baselines where the feature extractor was frozen. 
This subpar performance can be attributed to the specific nature of the WSI classification task. 
It requires processing all the patches in a given WSI during each iteration of model updating, which is computationally expensive. 
As a result, only a small portion of the instances are used for training, leading to a high risk of overfitting.
In contrast, our proposed method of using visual prompts allows for the quick learning of the policy of the current task based on the previously learned representation, enabling an efficient task and domain adaption, thereby overcoming the limitations of fine-tuning. 

%% file: Table/table1.tex
\begin{table}[t!]
\caption{Results (\%) of comparative experiments on Camelyon-16 and TCGA-NSCLC dataset. In the table, the best results are in bold. “RPS”: representative patch selection, “FT”: fine-tuning, “PT”: prompt fine-tuning.}
\setlength{\tabcolsep}{1.5mm}{

\begin{tabular}{c|l|ccc|ccc}
\hline
\multirow{2}{*}{Dataset}         & \multicolumn{1}{c|}{\multirow{2}{*}{Method}} & \multicolumn{3}{c|}{ResNet-18} & \multicolumn{3}{c}{ResNet-50} \\
                                 & \multicolumn{1}{c|}{}    & AUC   & F1    & Acc   & AUC   & F1    & Acc     \\ 
                                 \hline\hline
\multirow{6}{*}{Camelyon16}      & DTFD \cite{DTFD}        & 87.11 & 78.05 & 86.05 & 90.07 & 81.63 & 86.05   \\
                                 & DTFD-RPS-FT             & 81.40 & 69.23 & 81.58 & 89.34 & 81.32 & 86.82   \\
                                 & \cellcolor{light_cyan}DTFD-RPS-PT & \cellcolor{light_cyan}\textbf{90.94} & \cellcolor{light_cyan}\textbf{78.72} & \cellcolor{light_cyan}\textbf{84.50} & \cellcolor{light_cyan}\textbf{91.40} & \cellcolor{light_cyan}\textbf{83.52} & \cellcolor{light_cyan}\textbf{88.37} \\ 
                                 \cline{2-8} 
                                 & ABMIL \cite{ABMIL}              & 84.08 & 74.07 & \textbf{83.72} & 85.96 & 77.11 & 85.27   \\
                                 & ABMIL-RPS-FT            & 83.98 & 76.19 & 82.95 & 83.78 & \textbf{80.90} & \textbf{86.82}    \\
                                 & \cellcolor{light_cyan}ABMIL-RPS-PT            &\cellcolor{light_cyan}\textbf{87.45} & \cellcolor{light_cyan}\textbf{76.40} & \cellcolor{light_cyan}\textbf{83.72} & \cellcolor{light_cyan}\textbf{86.73} & \cellcolor{light_cyan}78.65 & \cellcolor{light_cyan}85.27  \\ 
                                  \hline\hline
\multirow{6}{*}{\begin{tabular}[c]{@{}c@{}}TCGA\\ NSCLC\end{tabular}} 
                                 & DTFD \cite{DTFD}           & 92.91 & \textbf{87.22} & \multicolumn{1}{c|}{87.12} & 93.47 & 89.38 & 89.01 \\
                                 & DTFD-RPS-FT             & 92.97 & 85.28 & \multicolumn{1}{c|}{85.22}  &  92.27 &  86.55  & 85.99     \\
                                 & \cellcolor{light_cyan}DTFD-RPS-PT             &\cellcolor{light_cyan}\textbf{94.19} & \cellcolor{light_cyan}87.16& \multicolumn{1}{c|}{\cellcolor{light_cyan}\textbf{87.50}}&\cellcolor{light_cyan}\textbf{93.80}&\cellcolor{light_cyan}\textbf{89.71} & \cellcolor{light_cyan}\textbf{89.39} \\ \cline{2-8} 
                                 & ABMIL \cite{ABMIL}            & 93.56        &  86.05       &  86.36        &    93.34      &  88.39       & 88.26        \\
                                 & ABMIL-RPS-FT            & 89.76        &  83.22       &  81.06        &   90.85      &	85.29         &   	84.85      \\
                                 & \cellcolor{light_cyan}ABMIL-RPS-PT            & \cellcolor{light_cyan}\textbf{93.84} &  \cellcolor{light_cyan}\textbf{87.46}   & \cellcolor{light_cyan}\textbf{86.74}  &  \cellcolor{light_cyan}\textbf{94.21}  & \cellcolor{light_cyan}\textbf{89.55}   &  \cellcolor{light_cyan}\textbf{89.39}       \\ 
                                 \hline
\end{tabular}}
\label{table1}%
\end{table}

%% file: Figure/Ablation.tex
\begin{figure*}[t!]
	\centering
	\includegraphics[width=1\textwidth]{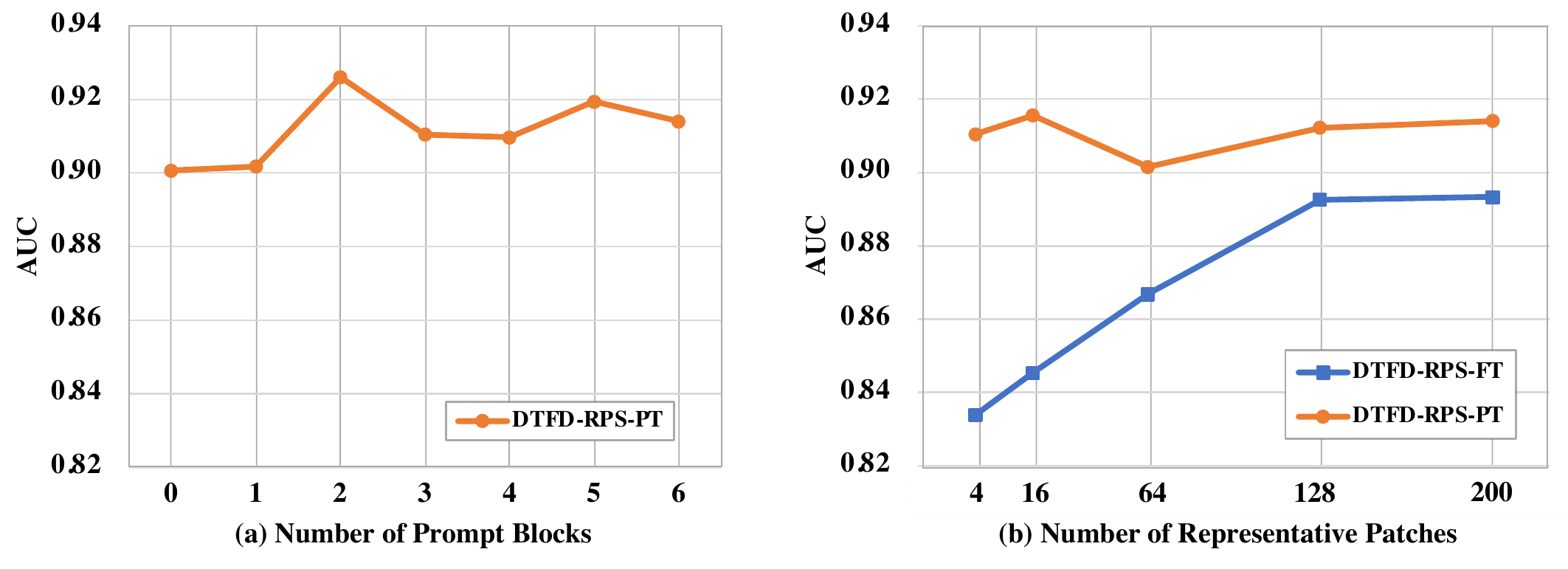}
	\caption{Results of ablation studies on CAMELYON16 dataset: (a) Effectiveness of the number of prompt blocks (b) Influence of the number of representative patches.}
	\label{fig_ablation}
\end{figure*}

%% file: Section/conclusion.tex
\section{Conclusion}
In this paper, we propose a novel prompt learning method to learn domain-specific knowledge transformation from ImageNet pre-trained models to pathological images. 
Our innovation is based on the observation that there is a large domain shift and task discrepancy between the upstream datasets and pathological tasks, resulting in sub-optimal feature representation.
To relieve this issue, we introduce a prompt component and representative patches selection strategy to fine-tune the prompt blocks while freezing the feature extractor backbone.
In this way, the extracted patch features can be adapted for pathological images and boost the WSI classification with MIL models.
Experiments on two public datasets (\ie, Cameylon16 and TCGA-NSCLC) with two MIL classifiers (\ie, DTFD and ABMIL) demonstrate the effectiveness and efficiency of our method. 